\documentclass{article}

     \usepackage[final,nonatbib]{neurips_2024}

\usepackage[utf8]{inputenc} 
\usepackage[T1]{fontenc}    
\usepackage{hyperref}       
\usepackage{url}            
\usepackage{booktabs}       
\usepackage{amsfonts}       
\usepackage{nicefrac}       
\usepackage{microtype}      
\usepackage{graphicx}
\usepackage{cancel}
\usepackage{amsmath}
\usepackage{wrapfig}
\usepackage{threeparttable}
\usepackage{multirow}
\usepackage[table]{xcolor}
\usepackage{booktabs}
\usepackage{arydshln}
\usepackage{float}
\usepackage{subfig}
\usepackage{makecell}

\definecolor{lightgray}{gray}{0.8}
\definecolor{lightblue}{rgb}{0.21,0.49,0.74}
\hypersetup{
    colorlinks=true,
    breaklinks=true,
    urlcolor=lightblue,
    linkcolor=red,
    bookmarksopen=false,
    filecolor=black,
    citecolor=lightblue,
    linkbordercolor=lightblue
}

\title{Cross-video Identity Correlating for Person Re-identification Pre-training}

\author{%
	\textbf{Jialong Zuo $^{1}$} \quad ~\textbf{Ying Nie $^2$} \quad ~\textbf{Hanyu Zhou $^1$} \quad ~\textbf{Huaxin Zhang $^1$} \\ ~\textbf{Haoyu Wang $^2$} \quad ~\textbf{Tianyu Guo $^{2}$} \quad ~\textbf{Nong Sang $^1$} \quad ~\textbf{Changxin Gao $^1$}\thanks{Corresponding Author. Project Link: \url{https://github.com/Zplusdragon/CION_ReIDZoo}}\\
    \\
	$^{1}$ National Key Laboratory of Multispectral Information Intelligent Processing Technology,\\ School of Artificial Intelligence and Automation, Huazhong University of Science and Technology, \\ $^{2}$ Huawei Noah’s Ark Lab.   \\
    {\tt\small\{jlongzuo, cgao\}@hust.edu.cn}
}

\begin{document}

\maketitle

\vspace{-8mm}
\begin{abstract}
\vspace{-3mm}
Recent researches have proven that pre-training on large-scale person images extracted from internet videos is an effective way in learning better representations for person re-identification. However, these researches are mostly confined to pre-training at the instance-level or single-video tracklet-level. They ignore the identity-invariance in images of the same person across different videos, which is a key focus in person re-identification. To address this issue, we propose a Cross-video Identity-cOrrelating pre-traiNing (CION) framework. Defining a noise concept that comprehensively considers both intra-identity consistency and inter-identity discrimination, CION seeks the identity correlation from cross-video images by modeling it as a progressive multi-level denoising problem. Furthermore, an identity-guided self-distillation loss is proposed to implement better large-scale pre-training by mining the identity-invariance within person images. We conduct extensive experiments to verify the superiority of our CION in terms of efficiency and performance. CION achieves significantly leading performance with even fewer training samples. For example, compared with the previous state-of-the-art~\cite{ISR}, CION with the same ResNet50-IBN achieves higher mAP of 93.3\% and 74.3\% on Market1501 and MSMT17, while only utilizing 8\% training samples. Finally, with CION demonstrating superior model-agnostic ability, we contribute a model zoo named ReIDZoo to meet diverse research and application needs in this field. It contains a series of CION pre-trained models with spanning structures and parameters, totaling 32 models with 10 different structures, including GhostNet, ConvNext, RepViT, FastViT and so on.
\end{abstract}

\vspace{-6mm}
\section{Introduction}
\vspace{-3mm}
Person re-identification (ReID), which aims to identify and match a target person across different camera views, is becoming increasingly influential in widespread applications, such as criminal tracking and missing individual searching. Although current ReID methods~\cite{MGN,ABDNet,SCWM,SAN,Nformer} have achieved significant progress for the past few years, it has become evident through recent research advancements that the mere development of sophisticated specialist algorithms has already encountered a performance bottleneck. 

\begin{figure}[htb]
\centering
\includegraphics[width=\linewidth]{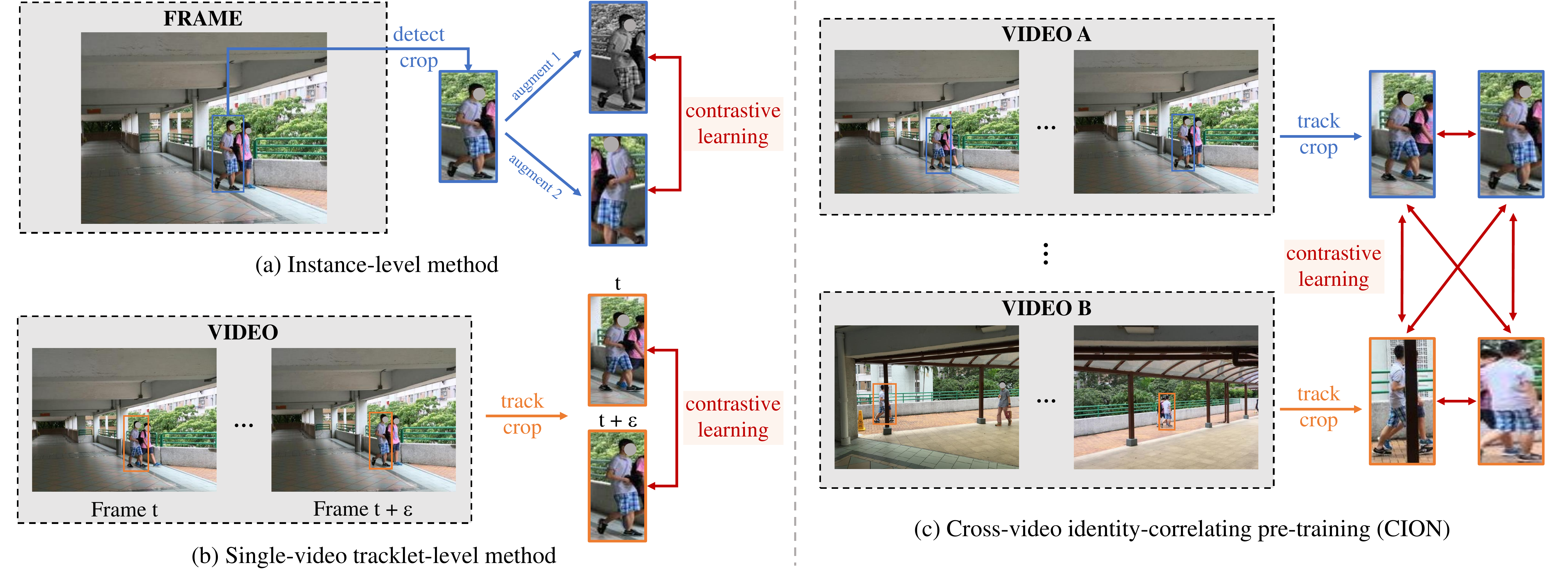}
\caption{Comparisons between our proposed CION with other pre-training methods. In (a), the instance-level method mines \textit{instance-invariance} by contrastive learning on augmented views of each image, completely ignoring the invariance within different images of the same person. In (b), the single-video tracklet-level method mines \textit{tracklet-invariance} by contrastive learning on images of each tracklet in single video, significantly ignoring the invariance in images of the same person across different videos. In (c), our CION learns \textit{identity-invariance} by correlating the images of the same person across different videos, thus leading to better representation learning.
}
\vspace{-10mm}
\label{fig:introduction}
\end{figure}

Meanwhile, pre-training on large-scale images~\cite{DINO,MoCov3,MAE,CLIP} has shown great success to further improve model performance in the field of general vision. However, due to the significant domain gap, these pre-trained models can only provide very limited improvement for person re-identification. Therefore, with the vast amount of person-containing videos available on the internet, some researches~\cite{LUP,LUPNL,PASS,ISR,UPReID,TranSSL} turn to exploring the potential of pre-training on person images extracted from such videos and demonstrate encouraging results. However, due to the unavailability of identity labels for such extracted person images, these pre-training methods are confined to learning representations at the instance-level or single-video tracklet-level, as shown in Fig.~\ref{fig:introduction}

On the one hand, the instance-level methods~\cite{LUP,PASS,UPReID,TranSSL} completely overlook the identity-invariance in different images of the same person. They incorporate some person-related prior knowledge into general pre-training frameworks, neglecting the concept of person identity. For example, LUP~\cite{LUP} improves MoCoV2~\cite{MoCOV2} by systematically studying the data augmentation and contrastive loss, while PASS~\cite{PASS} improves DINO~\cite{DINO} by generating part-level features to offer fine-grained information. However, there exists a notable concordance within different images of the same person. Directly ignoring such a concordance easily leads to learning weaker person representations.

On the other hand, the single-video tracklet-level methods~\cite{LUPNL,ISR} only exploit the short-range tracklet-invariance while significantly neglecting the identity-invariance in images of the same person across different videos. They simply regard one tracklet in a single video as one person, which will result in two substantial problems. 1) Insufficient intra-identity consistency. Due to the unavoidable inaccurate tracking, there are quite a few images within a tracklet that do not belong to the same person. Roughly considering these images as coming from the same person will lead to insufficient intra-identity consistency. 2) Limited inter-identity discrimination. It is quite evident that there are considerable instances of the same person appearing in different videos, thus being treated as different tracklets (identities). Roughly considering these images as coming from different persons will lead to limited inter-identity discrimination.

To address these limitations mentioned above, as shown in Fig.~\ref{fig:introduction}, we propose a Cross-video Identity-cOrrelating pre-traiNing (CION) framework, by which the images of the same person across different videos are explicitly correlated to deeply mine the identity-invariance. CION first seeks the identity correlation from long-range cross-video images by modeling it as a progressive multi-level denoising problem, where a noise concept that comprehensively considers both intra-identity consistency and inter-identity discrimination is defined. Subsequently, with the sought identity correlation, CION introduces an identity-guided self-distillation loss to implement better pre-training by learning the invariance within augmented views in multiple images of the same person. 

We conduct extensive experiments to verify the superiority of our CION in terms of efficiency and performance. CION achieves the best performance with significantly fewer training samples. With the ResNet50-IBN backbone, compared with the previous state-of-the-art method ISR~\cite{ISR}, our CION achieves higher mAP of 93.3\% and 74.3\% on the Market1501 and MSMT17 datasets, while only utilizing 8\% training samples. Noting that ISR is pre-training at the single-video tracklet-level, this experimental result fully validates the necessity of learning identity-invariance across different videos. 

Finally, we contribute a fully open-sourced model zoo named ReIDZoo to meet diverse research and application needs within this field. With CION demonstrating superior performance improvements to diverse model structures (model-agnostic ability), we pre-trained a series of models with spanning structures and parameters, totaling 32 models with 10 different structures, including GhostNet~\cite{GhostNet}, EdgeNext~\cite{EdgeNext}, ConvNext~\cite{ConvNext}, RepViT~\cite{RepViT}, FastViT~\cite{FastViT} and so on.

In conclusion, the highlights of our work can be summarized into three points:

(1)	We propose a Cross-video Identity-cOrrelating pre-traiNing (CION) framework. It explicitly mines the identity-invariance in person images extracted from internet videos by progressive multi-level denoising and identity-guided self-distillation, leading to better representation learning. 

(2)	Extensive experiments verify the superiority of CION in terms of efficiency and performance. It achieves the best performance with much fewer training samples compared with previous methods, while also demonstrating adaptability to diverse model structures with spanning parameters.

(3)	We contribute a model zoo named ReIDZoo. It contains a series of CION pre-trained models with spanning structures and parameters, totaling 32 models with 10 different structures. It conveniently meets various research and application needs, and greatly promotes the development of this field.

\section{Related work}
\textbf{Self-supervised pre-training for general vision.} 
In recent years, self-supervised pre-training methods have shown promising results in learning representation from large-scale data. 
MoCo~\cite{MoCOV2} facilitates contrastive learning by building a dynamic dictionary with a queue and a moving-averaged encoder.
BYOL~\cite{BYOL} relies on two networks, referred to as online and target networks, that interact and learn from each other with a slow-moving average strategy.
DINO~\cite{DINO} proposes a self-distillation framework by underlying the importance of momentum encoder, multi-crop training and so on.
MAE~\cite{MAE} adopts the ViT backbone and learns representations by predicting the masked patches from the remaining visible ones.
However, due to significant domain gap and neglect of person-related characteristics, these general methods show poor transferability to person re-identification.

\textbf{Self-supervised pre-training for person re-identification.} 
Existing pre-training methods can be classified into two categories, the first being the instance-level methods~\cite{LUP,PASS,UPReID,TranSSL,SOLIDER} and the second being single-video tracklet-level method~\cite{ISR,LUPNL}.
The instance-level methods strive to merge the person-related prior knowledge into existing self-supervised learning methods. For example, LUP~\cite{LUP} improves MoCov2~\cite{MoCOV2} through systematical studying of data augmentation and contrastive loss. PASS~\cite{PASS} improves DINO~\cite{DINO} by generating part-level features to offer fine-grained information. SOLIDER~\cite{SOLIDER} turns to utilize pseudo semantic label and import some semantic information into the learned representations. However, these instance-level methods are incapable of harnessing the identity-invariance in different images of the same person.
Meanwhile, the single-video tracklet-level methods strive to mine some invariance by regarding one tracklet in a single video as one person. For example, LUP-NL~\cite{LUP} utilizes supervised ReID learning, label-guided contrastive learning and prototype-based contrastive learning to rectify the noisy tracklet-level labels in person images extracted from raw internet videos. ISR~\cite{ISR} leverages massive data to construct the positive pairs from inter-frame images in a single videos. The identity-invarience of the same person across different videos is significantly neglected by these single-video tracklet-level methods.

\section{CION: Cross-video Identity-cOrrelating pre-traiNing}
We present a cross-video identity-correlating pre-training (CION) framework, which explicitly learns identity-invariance in long-range cross-video person images. First, we define a noise concept that comprehensively considers both intra-identity consistency and inter-identity discrimination (Sec.~\ref{sec3.1}). Next, we model the identity correlation seeking process from cross-video images as a progressive multi-level denoising problem (Sec.~\ref{sec3.2}). Finally, we propose an identity-guided self-distillation loss to mine identity-invariance from person images with the sought identity correlation (Sec.~\ref{sec3.3}).

\subsection{Noise definition}
\label{sec3.1}
We demonstrate that a sample set with identity labels should have the following characteristics. 
1) Intra-identity consistency. Samples belonging to the same identity should aggregate around a centroid in a high-dimensional space, i.e., they should be enveloped by a hypersphere $s_k = (\mathbf{c}_k,r_k)$, where $\mathbf{c}_k$ represents the centroid of the sphere and $r_k$ represents its radius. A smaller radius of the sphere indicates a denser degree of aggregation, which signifies higher intra-identity consistency.
2) Inter-identity discrimination. Samples belonging to two different identities should be enveloped by two distinct hyperspheres $s_i$ and $s_j$ respectively, with a certain distance $d_{i,j}$ maintained between the centroids of $\mathbf{c}_i$ and $\mathbf{c}_j$. A larger distance indicates a sparser distribution of hyperspheres, implying higher intra-identity discrimination.
In summary, the smaller radii of the hyperspheres and the greater distances between the hypersphere centroids mean more desirable intra-identity consistency and inter-identity discrimination.

Considering the above demonstration, for a sample set $S = \{\mathbf{x}_i^{y_i}\}_{i=1}^{N}$, where the identity label $y_i \in \{1,2,3,…,M\}$, we can further regard $S$ as a hypersphere set $\{s_{k}\}_{k=1}^M$, where $\mathbf{x}_i \in s_k$ if $y_i=k$. Then for a single hypersphere $s_k$, the intra-consistency noise is formulated as:

\begin{equation}\label{intra-consistency noise}
    n_k^{cst}= \mu(r_k-\sigma_{cst}) \quad s.t. \quad r_k=\max_i(dist(\mathbf{x}_i,\mathbf{c}_k)) \quad \forall \mathbf{x}_i \in s_k,
\end{equation}
where $\mu(\cdot)$ is an unit step function, $dist(\cdot)$ is any distance metric such as cosine distance, and the centroid $\mathbf{c}_k$ is calculated by averaging all samples in $s_k$. We can observe that this formula determines the presence of noise by explicitly constraining the maximum radius of the hypersphere, indicating whether the aggregation degree of samples with the same identity meets the required criteria $\sigma_{cst}$.

Meanwhile, for two distinctive hyperspheres $s_i$ and $s_j$, the inter-discrimination noise between them is formulated as:

\begin{equation}\label{inter-discrimination noise}
    n_{i,j}^{drm}= \mu(\sigma_{drm}-d_{i,j}) \quad s.t. \quad d_{i,j} = dist(\mathbf{c}_i,\mathbf{c}_j).
\end{equation}
We can observe that this formula explicitly constrains the minimum distance between the centroids of the hyperspheres to determine the presence of noise, indicating whether the margin between samples of different identities meets the required criteria $\sigma_{drm}$. 

Finally, comprehensively considering both intra-identity consistency and inter-identity discrimination, the overall noise for a sample set $S = \{s_{k}\}_{k=1}^M$ can be formulated as:

\begin{equation}\label{overall noise}
    n_{S}= \frac{1}{M}(\sum_{k=1}^{M}{n_k^{cst}} + \sum_{i=1}^{M}\sum_{j=1,j\neq{i}}^{M}{n_{i,j}^{drm}}).
\end{equation}
We consider that a sample set with a favorable level of identity correlation should exhibit the smallest possible amount of $n_{S}$ in order to achieve superior consistency and discrimination.

\subsection{Progressive multi-level denoising}
\label{sec3.2}
Given crawled internet videos, a straightforward approach~\cite{LUPNL,ISR} is to utilize tracking algorithms to extract person tracklets from each video and treat each tracklet as an unique identity, thereby forming a sample set with initial identity correlation. 
However, due to inevitable false positives in tracking and the neglect of the same person across different videos, the identity correlation in this sample set is significantly noisy, lacking satisfactory intra-identity consisitency and inter-identity discrimination.
Therefore, we propose a progressive multi-level denoising strategy to seek superior identity correlation by minimizing the overall $n_{S}$ of the sample set as much as possible.

\textbf{\textit{Single-tracklet denosing.}}
First, we denoise each single tracklet to guarantee the intra-identity consistency. For an initial single-tracklet sample set with $m$ extracted image features $t^{0} = \{\mathbf{x}_i\}_{i=1}^m$, we obtain the denoised sample set by constraining the maximum hypersphere radius through multiple iterations.
That is, for the sample set $t^{j}$ of the $j$-th iteration, we compute the distances between each sample and the centroid of the remaining samples to identify the sample with the maximum deviation. If the deviation exceeds $\sigma_{cst}$, we remove the sample from $t^{j}$ to an exclusion set $t_{exc}$ and proceed to the next iteration; otherwise, we terminate the iteration and regard $t^{j}$ as the denoised sample set with good intra-identity consistency. The iterative process can be formulated as:
\begin{equation}
\begin{cases}
\mathbf{x}_{dev}^j = \arg\max_{\mathbf{x}_i}dist(\mathbf{x}_i,\mathbf{c}_{t^j\backslash{\mathbf{x}_i}}) \quad \forall \mathbf{x}_i \in t^{j}\\
t^{j+1} = t^{j}\backslash{\mathbf{x}_{dev}^j}, \quad t_{exc} = t_{exc} \cup \mathbf{x}_{dev}^j \quad s.t. \quad dist(\mathbf{x}_{dev}^j,\mathbf{c}_{t^j\backslash{\mathbf{x}_{dev}^j}}) \ge \sigma_{cst},
\end{cases} \label{single-tracklet denosing formular}
\end{equation}
We can observe that by iteratively constraining the maximum hypersphere radius of the tracklet, we achieve the exclusion of anomalous samples. In essence, this iterative process aims to eliminate the noise $n^{cst}$ as defined in Eq.~\ref{intra-consistency noise}, thereby ensuring favorable intra-identity consistency for each tracklet.

\begin{figure}[htb]
\centering
\includegraphics[width=\linewidth]{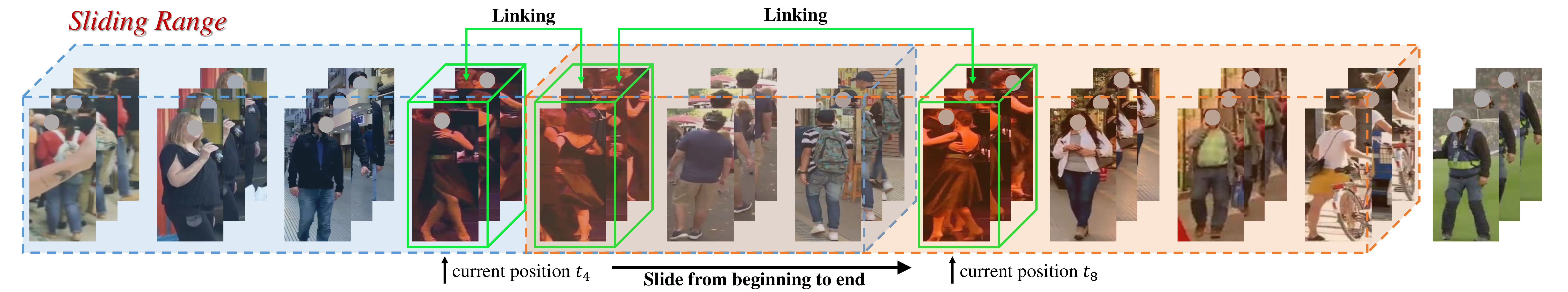}
\caption{A toy example for Sliding Range and Linking Relation.}
\vspace{-5mm}
\label{fig:sliding_range}
\end{figure}

\textbf{\textit{Short-range single-video denosing.}} Then, we denoise each single video with multiple tracklets to guarantee both intra-identity consistency and inter-identity discrimination. For an initial single-video sample set with $n$ denoised tracklets $v = \{t_k\}_{k=1}^{n}$and respective exclusion sets $\{t_{exc,k}\}_{k=1}^{n}$ , we first reallocate the anomalous samples of each exclusion set to ensure their proper tracklet assignment. The reallocation process for an anomalous sample $\mathbf{x}_i^{k} \in t_{exc,k}$ can be formulated as:
\begin{equation}
\begin{cases} \label{reallocation process}
t_{min} = \arg \min_{t_r}dist(\mathbf{x}_i^{k},\mathbf{c}_{t_r})  \quad \forall t_r \in v\backslash{t_k} \\
  t_{min} = t_{min} \cup \mathbf{x}_i^{k} \quad s.t. \quad dist(\mathbf{x}_i^{k},\mathbf{c}_{t_{min}})<\sigma_{cst}.
\end{cases} 
\end{equation}
Through the aforementioned reallocation process, we can assign each anomalous sample to its correct tracklet, or discard those extreme anomalous samples that do not belong to any tracklet, such as samples devoid of person presence. Subsequently, we merge tracklets with closely located hypersphere centroids to ensure robust inter-identity discrimination, which can be formulated as:
\begin{equation}
\label{merging process}
merge(t_p,t_q) = \mu(\sigma_{drm}-dist(\mathbf{c}_{t_p},\mathbf{c}_{t_q})) \quad s.t. \quad \forall t_p ,t_q \in v,
\end{equation}
where $\mu(\cdot)$ is an unit step function to indicate whether the merge operation is applied to $t_p$ and $t_q$. 

We can observe that through the aforementioned two-stage process, we achieve the reallocation of anomalous samples and the merging of similar tracklets. In essence, this two-stage process aims to eliminate the overall noise $n_S$ as defined in Eq.~\ref{overall noise}, thereby ensuring both favorable intra-identity consistency and inter-identity discrimination for each video.

\textbf{\textit{Long-range cross-video denosing.}} Finally, we denoise multiple videos to seek identity correlation across videos. A straightforward method for denoising a large-scale sample set containing many videos is to concatenate the videos together and treat them as a single ultra-long video, then apply the single-video denoising strategy mentioned above. However, when a video contains $N$ tracklets, the computational complexity of Eq.~\ref{merging process} is $\mathcal{O}(N^2)$. For an ultra-long video with an overwhelming number of tracklets, this undoubtedly introduces unmanageable computational overhead. To address this issue, we design Sliding Range and Linking Relation to implement ultra-long video denoising.

As shown in Fig.~\ref{fig:sliding_range}, for an ultra-long concatenated video composed of $N$ sub-video tracklets $V = \{t_i\}_{i=1}^N$, we construct a Sliding Range $SR_c=\{c,r_{s}\}_{c:1\to N}$ to slide from the beginning to the end of $V$, where $c$ represents the index of the current position, and $r_s$ represents the half-width of $SR_c$. Then, for all tracklets covered by $SR_c$, we obtain the Linking Relation between each tracklet and $t_c$ by utilizing the merging function defined in Eq.~\ref{merging process}, which is formulated as:  
\begin{equation}
\label{relation link process}
LR(t_i,t_c) = merge(t_i,t_c)\quad s.t. \quad \forall t_i\in SR_c\backslash{t_c},
\end{equation}
where $LR(\cdot)$ indicates whether a bidirectional Linking Relation will be established between the two tracklets. Subsequently, after the Sliding Range traverses the entire video, we perform a closure operation based on the obtained Linking Relations to derive the closures. Finally, the tracklets within each closure are merged as one single identity.

We can observe that the designing of Sliding Range significantly reduces the computational complexity from $\mathcal{O}(N^2)$ to $\mathcal{O}(N)$, which makes it possible to denoise massive-scale crawled videos. Meanwhile, the closure operation based on Linking Relations ensures long-range identity correlation for different tracklets of the same person across different videos. In essence, this long-range cross-video denosing strategy aims to eliminate the inter-discrimination noise $n^{drm}$ as defined in Eq.~\ref{inter-discrimination noise} at the multi-video level, thereby further ensuring favorable inter-identity discrimination of all videos.

\textbf{\textit{Identity correlation seeking.}}
We summarize the whole process of seeking the identity correlation from large-scale crawled person-containing videos as follows. Firstly, we should utilize an off-the-shelf person tracking algorithm to extract person tracklets from each video and assign each tracklet with a class label, forming a noisy sample set with initial identity correlation. Then, the noisy sample set should undergo single-tracklet denoising, short-range single-video denoising and long-range cross-video denoising in sequence to seek satisfactory identity correlation. Finally, we consider all images within a single denoised tracklet as belonging to the same person, and this denoised sample set can serve as a well-prepared dataset with good identity correlation for subsequent pre-training.

\subsection{Identity-guided self-distillation}
\label{sec3.3}
\begin{wrapfigure}[27]{r}{6cm}
\vspace{-5mm}
\centering
\includegraphics[width=\linewidth]{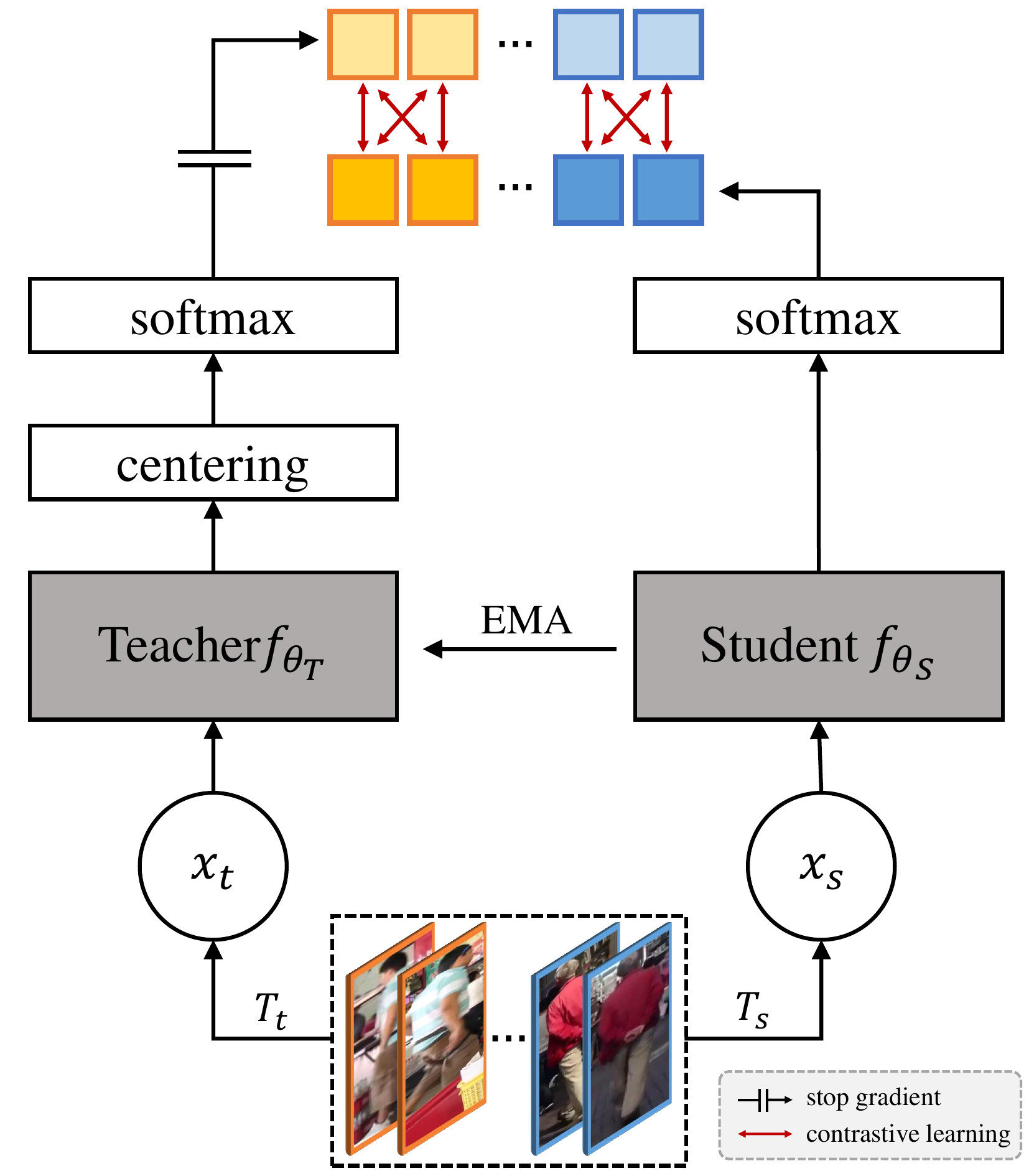}
\caption{Self-distillation with identity guidance. The overall structure shares a similarity with~\cite{DINO}, while the concept of identity is introduced. We illustrate it in the case of $N_{id}=2$ and pairs of views $(\mathbf{x}_t,\mathbf{x}_s)$ for simplicity. $T_t$ and $T_s$ represent different random transformations. All transformed views from images of the same person will engage in contrastive learning. }
\label{fig:IDINO}
\end{wrapfigure}
With the sought identity correlation, we propose to utilize simple but effective identity-guided self-distillation to pre-train our models, leading to better identity-invariance learning. The whole framework shares a similar overall structure as the popular self-distillation pre-training paradigm DINO for general vision~\cite{DINO}, as illustrated in Fig.~\ref{fig:IDINO}. Differently, we introduce the concept of identity and conduct contrastive learning on the augmented views from images of the same person. We train a student network $f_{\theta_S}$ to match the output probability distribution of a teacher network $f_{\theta_T}$, in which the two share the same architecture and are parameterized by $\theta_S$ and $\theta_T$, respectively.
Given an input image $\mathbf{x}$, both networks predict probability distributions over $K$ dimensions denoted by $P_S$ and $P_T$, which are obtained by normalizing the output of each network with a softmax function and formulated as:
\begin{equation}
\label{dino}
P_{S}(\mathbf{x})^{(i)}=\frac{\exp \left(f_{\theta_{S}}(\mathbf{x})^{(i)} / \tau_{s}\right)}{\sum_{k=1}^{K} \exp \left(f_{\theta_{S}}(\mathbf{x})^{(k)} / \tau_{s}\right)},
\end{equation}
where $P_{S}(\mathbf{x})$ is the predicted distribution of the image $\mathbf{x}$ for $f_{\theta_{S}}$ and $\tau_{s}>0$ is a hyperparameter controlling the output distribution sharpness. The similar formular holds for $P_T$ with $\tau_{t}$. 

In the following, we detail how we introduce the identity concept into self-distillation. First, with the sought identity correlation, we construct an identity set containing $N_{id}$ images of the same person $X = \{\mathbf{x}_{k}\}_{k=1}^{N_{id}}$. Then, given an image $\mathbf{x}_k$, we construct a set $V_k$ of different augmented views. This set contains two global views $\mathbf{x}_k^{g,1}$ and $\mathbf{x}_k^{g,2}$ and several local views of smaller resolution. We pass the whole set $V_k$ through the student while only pass the global views through the teacher. Finally, the student learns to match the output probability distribution of teacher by minimizing the cross-entropy:
\begin{equation}
\label{dino_ce}
\min _{\theta_{S}} \sum_{\mathbf{x}_i \in X} \sum_{\mathbf{x}_j \in X} \sum_{\mathbf{x} \in\left\{\mathbf{x}_{i}^{g,1}, \mathbf{x}_{i}^{g,2}\right\}} \sum_{\substack{\mathbf{x}^{\prime} \in V_j \\ \mathbf{x}^{\prime} \neq \mathbf{x}}} H\left(P_{T}(\mathbf{x}), P_{S}\left(\mathbf{x}^{\prime}\right)\right),
\end{equation}
where $H(a,b)=-a\log{b}$ and the parameters $\theta_{S}$ are optimized with stochastic gradient descent. Following~\cite{DINO}, the teacher network is frozen over an epoch and updated with an exponential moving average (EMA) on the student weights. Meanwhile, the output of the teacher network is centered with a mean calculated over the batch.

\section{Experiments}
\subsection{Experimental setup}
\label{setup}

\textbf{Pre-training and fine-tuning.} During pre-training, we directly regard the images of SYNTH-PEDES~\cite{PLIP} as a noisy sample set with initial identity correlation. Then, we utilize our proposed progressive multi-level denoising strategy to seek the identity correlation in it. Meanwhile, to ensure a fair comparison with previous methods, we adopt random selection to align the overall sample size with that of the commonly used pre-training dataset LUPerson~\cite{LUP}. We name the whole sample set as CION-AL, and it contains 3,898,086 images of 246,904 identities totally. Finally, we utilize the CION-AL and our proposed identity-guided self-distillation loss to pre-train our models. We pre-trained 32 models of 10 architectures in total. During fine-tuning, as a common practice, we directly utilize TransReID~\cite{TransReID} to fine-tune our pre-trained models with input size as $256\times128$ for the supervised person ReID setting, if not specified. Also, the commonly used C-Contrast~\cite{CContrast} is adopted for the settings of unsupervised domain adaptation (UDA) and unsupervised learning (USL).

\textbf{Datasets and evaluation metrics.}
Following common practices, we conduct experiments on two datasets, \textit{i.e.}, Market1501~\cite{Market1501} and MSMT17~\cite{MSMT17}, for evaluating the performance on ReID tasks. Market1501 and MSMT17 are widely used person ReID datasets, which contain 32,688 images of 1,501 persons and 126,411 images of 4,101 persons, respectively. We use the cumulative matching characteristics at top-1 (Rank-1) and mean average precision (mAP) as the evaluation metrics.

\textbf{Implementation details.}
To implement the progressive multi-level denoising, we utilize a ViT-Base model~\cite{TranSSL} trained on the images of UFine6926~\cite{UFineBench} as a normalized feature extractor. The intra-consisitency criteria $\sigma_{cst}$ and inter-discrimination criteria $\sigma_{drm}$ are set as 0.2 and 0.18, respectively. We utilize cosine distance to calculate the distance between two samples. The half-width $r_s$ of the sliding range is set as 1000. For pre-training, the global views and local views are resized to $256\times128$ and $128\times64$, respectively. We pre-train our models on $8\times$V100 GPUs for 100 epochs. We adopt a curriculum learning strategy for setting the images per identity, where the initial $N_{id}$ is set to 2, and at epochs 40, 60, and 80, $N_{id}$ increases to 4, 6, and 8, respectively. To optimize the utilization of GPU memory, the batch size per GPU and the number of cropped local views are adjusted according to the varying parameter counts of the models. All other settings for pre-training, such as learning rate, are consistent with those used in DINO~\cite{DINO}. For fine-tuning, we employ the commonly used  methods~\cite{TransReID,MGN,CContrast} and make only slight adjustments to settings such as learning rate and batch size. Specific settings and more details can be found in Sec.~\ref{pretrainingsetting} of the appendix.

\vspace{-2mm}
\subsection{Comparison with state-of-the-art methods}
\begin{wraptable}[32]{r}{7.8cm}
\vspace{-6mm}
\small
\tabcolsep=0.5pt
\begin{center}
\caption{Comparison with SoTA methods of supervised person ReID. * means the result with the input size as $384\times128$. $^{\dag}$ represents fine-tuning with MGN.
}
\label{tab:supervised_reid}
\begin{threeparttable}
\begin{tabular}{lccccc}
\toprule[1pt]
\multirow{2}{*}{\textbf{Method}} & \multirow{2}{*}{\textbf{Backbone}} & \multicolumn{2}{c}{\textbf{Market1501}}  & \multicolumn{2}{c}{\textbf{MSMT17}}  \\ \cline{3-6}
       &  & \textbf{mAP} & \textbf{Rank1} & \textbf{mAP} & \textbf{Rank1} \\ 
       \hline 
\multicolumn{6}{l}{\textit{Pre-training on ImageNet1K-1.3M}} \\
MGN$^*$~\cite{MGN} & R50 & 87.5 & 95.1 & 63.7 & 85.1  \\
ABDNet$^*$~\cite{ABDNet} & R50 & 88.3 & 95.6 & 60.8 & 82.3 \\
SAN~\cite{SAN} & R50 & 88.0 & 96.1 & 55.7 & 79.2 \\
NFormer~\cite{Nformer} & R50 & 91.1 & 94.7 & 59.8 & 77.3 \\
\hdashline[2.5pt/5pt]
\multicolumn{6}{l}{\textit{Pre-training on ImageNet21K-14M}} \\
TransReID~\cite{TransReID} & ViT-B & 87.4 & 94.6 & 63.6 & 82.5 \\
DCAL~\cite{DCAL} & ViT-B & 87.5 & 94.7 & 64.0 & 83.1 \\
\hline
\multicolumn{6}{l}{\textit{Pre-training with Large-scale Person Images}} \\
LUP$^{\dag*}$ (\textit{4.2M})~\cite{LUP} & R50 & 91.0 & 96.4 & 65.7 & 85.5 \\
UPReID$^{\dag*}$ (\textit{4.2M})~\cite{UPReID} & R50 & 91.1 & 97.1 & 63.3 & 84.3 \\ 
LUP-NL$^{\dag*}$ (\textit{10.7M})~\cite{LUPNL} & R50 & 91.9 & 96.6 & 68.0 & 86.0 \\
PLIP$^{\dag*}$ (\textit{4.8M})~\cite{PLIP} & R50 &91.3 & 96.7 & 66.2 & 85.3 \\
ISR$^{\dag*}$ (\textit{47.8M})~\cite{ISR} & R50-IBN & 92.3 & 96.9 & 71.5 & 88.4 \\
\rowcolor{lightgray} 
Ours$^{\dag*}$ (\textit{3.9M}) & R50 & \textbf{92.3} & \textbf{97.1} & \textbf{70.1} & \textbf{87.8} \\
\rowcolor{lightgray} 
Ours$^{\dag*}$ (\textit{3.9M}) & R50-IBN & \textbf{93.3} & \textbf{97.3} & \textbf{74.3} & \textbf{89.8} \\

MoCov3 (\textit{4.2M})~\cite{MoCov3} & ViT-S & 82.2 & 92.1 & 47.4 & 70.3 \\
TranSSL (\textit{4.2M})~\cite{TranSSL} & ViT-S & 91.1 & 95.9 & 66.8 & 85.5 \\
PASS (\textit{4.2M})~\cite{PASS} & ViT-S & 92.2 & 96.3 & 69.1 & 86.5 \\
PASS (\textit{4.2M})~\cite{PASS} & ViT-B & 93.0 & 96.8 & 71.8 & 88.2 \\
PATH (\textit{11M})~\cite{PATH} & ViT-B & 89.5 & 95.8 & 69.1 &84.3 \\
\rowcolor{lightgray} 
Ours (\textit{3.9M}) & ViT-S & \textbf{92.8} & \textbf{96.5} & \textbf{70.3} & \textbf{86.9} \\
\rowcolor{lightgray} 
Ours (\textit{3.9M}) & ViT-B & \textbf{93.1} & \textbf{96.9} & \textbf{72.6} & \textbf{88.5} \\

SOLIDER$^*$ (\textit{4.2M})~\cite{SOLIDER} & Swin-T & 91.6 & 96.1 & 67.4 & 85.9 \\
SOLIDER$^*$ (\textit{4.2M})~\cite{SOLIDER} & Swin-S & 93.3 & 96.6 & 76.9 & 90.8 \\
\rowcolor{lightgray} 
Ours$^*$ (\textit{3.9M}) & Swin-T & \textbf{92.2} & \textbf{96.8} & \textbf{71.1} & \textbf{88.1} \\
\rowcolor{lightgray} 
Ours$^*$ (\textit{3.9M}) & Swin-S & \textbf{93.4} & \textbf{96.6} & \textbf{77.0} & \textbf{90.4} \\
\bottomrule[1pt]
\end{tabular}
\end{threeparttable}
\end{center}
\end{wraptable}
\textbf{Supervised person ReID.} 
We compare our CION with several outstanding state-of-the-art methods on supervised person ReID in Tab.~\ref{tab:supervised_reid}. We divide previous methods by model initialization, \textit{i.e.}, utilizing supervised ImageNet pre-training~\cite{ImageNet} and self-supervised large-scale person images pre-training as backbone initialization, respectively. 
Compared with existing best methods pre-trained on ImageNet, our method significantly outperforms them without any extra complex designs. For example, our ViT-B achieves remarkable performance gains over the previous best method DCAL~\cite{DCAL} on Market1501 and MSMT17 by 5.6\% and 8.6\% mAP, respectively. It should be noted that DCAL utilizes much more images for pre-training than ours (14M vs 3.9M). 
On the other hand, compared with existing best self-supervised methods pre-trained on large-scale person images, our method still shows significant superiority over them. It is worth mentioning that the single-video tracklet-level method ISR~\cite{ISR} utilizes a substantially larger number of person images for pre-training than ours (47.8M vs 3.9M). However, both fine-tuning with MGN~\cite{MGN}, our ResNet50-IBN obtains 93.3\% and 74.3\% mAP on Market1501 and MSMT17, surpassing ISR by 1.0\% and 2.8\%, respectively. This effectively demonstrates the necessity of learning identity invariance across different videos.

\begin{table}[t]
\vspace{-3mm}
\caption{Comparison with state-of-the-art methods of unsupervised person ReID under two different settings. ``Mar'' and ``MS'' denote Market1501 and MSMT17 respectively. * denotes adding CFS~\cite{TranSSL}.}
\vspace{1mm}
\label{table:unsupervised}
\centering
\begin{minipage}{\linewidth}
\subfloat[Comparison on UDA person ReID.
]{
\small
\tabcolsep=0.8pt
\centering
\begin{tabular}{lccccc}
\toprule[1pt]
\multirow{2}{*}{\textbf{Method}} & \multirow{2}{*}{\textbf{Backbone}} & \multicolumn{2}{c}{\textbf{MS$\rightarrow$Mar}}  & \multicolumn{2}{c}{\textbf{Mar$\rightarrow$MS}}  \\ \cline{3-6}
       &  & \textbf{mAP} & \textbf{Rank1} & \textbf{mAP} & \textbf{Rank1} \\ 
       \hline
\multicolumn{6}{l}{\textit{Pre-training on ImageNet1K-1.3M}} \\
DG-Net++~\cite{DGNet} & R50 & 64.6 & 83.1 & 22.1 & 48.4 \\
MMT~\cite{MMT} & R50 & 75.6 & 83.9 & 24.0 & 50.1 \\
SpCL~\cite{SpCL} & R50 & 77.5 & 89.7 & 26.8 & 53.7 \\
C-Contrast~\cite{CContrast}& R50 & 82.4 & 92.5 & 33.4 & 60.5 \\
MCRN~\cite{MCRN} & R50 & - & - & 32.8 & 64.4 \\ \hline
\multicolumn{6}{l}{\textit{Pre-training on Large-scale Person Images}} \\
LUP~\cite{LUP} & R50 & 85.1 & 94.4 & 28.3 & 53.8 \\
LUP~\cite{LUP} & R50-IBN & 86.9 & 84.6 & 42.6 & 69.1 \\
TranSSL~\cite{TranSSL} & ViT-S & 89.6 & 95.6 & 55.0 & 77.9 \\
PASS~\cite{PASS} & ViT-S & 90.2 & 95.8 & 49.1 & 72.7 \\
TranSSL$^*$~\cite{TranSSL} & ViT-S & 89.9 & 95.5 & 57.8 & 79.5 \\

\rowcolor{lightgray} 
Ours & R50 & \textbf{90.6} & \textbf{96.1} & \textbf{57.4} & \textbf{81.3} \\
\rowcolor{lightgray} 
Ours & R50-IBN & \textbf{92.1} & \textbf{96.5} & \textbf{62.7} & \textbf{84.3} \\
\rowcolor{lightgray} 
Ours & ViT-S & \textbf{90.4} & \textbf{95.6} & \textbf{58.0} & \textbf{79.9} \\
\bottomrule[1pt]
\end{tabular}

}
\hspace{0em}
\subfloat[Comparison on USL person ReID.
]{
\small
\tabcolsep=0.8pt
\centering
\begin{tabular}{lccccc}
\toprule[1pt]
\multirow{2}{*}{\textbf{Method}} & \multirow{2}{*}{\textbf{Backbone}} & \multicolumn{2}{c}{\textbf{Mar}}  & \multicolumn{2}{c}{\textbf{MS}}  \\ \cline{3-6}
       &  & \textbf{mAP} & \textbf{Rank1} & \textbf{mAP} & \textbf{Rank1} \\ 
       \hline
\multicolumn{6}{l}{\textit{Pre-training on ImageNet1K-1.3M}} \\
MMCL~\cite{MMCL} & R50 & 45.5 & 80.3 & 11.2 & 35.4 \\
HCT~\cite{HCT} & R50 & 56.4 & 80.0 & - & - \\
IICS~\cite{IICS} & R50 & 72.9 & 89.5 & 26.9 & 52.4 \\
C-Contrast~\cite{CContrast} & R50 & 82.6 & 93.0 & 33.1 & 63.3 \\
MCRN~\cite{MCRN} & R50 & 80.8 & 92.5 & 31.2 & 63.6 \\ \hline
\multicolumn{6}{l}{\textit{Pre-training on Large-scale Person Images}} \\
LUP~\cite{LUP} & R50 & 84.0 & 93.4 & 31.4 & 58.8 \\
LUP~\cite{LUP} & R50-IBN & 86.4 & 94.2 & 39.8 & 66.1 \\
TranSSL~\cite{TranSSL} & ViT-S & 89.3 & 94.8 & 48.8 & 74.4 \\
PASS~\cite{PASS} & ViT-S & 88.5 & 94.9 & 41.0 & 67.0 \\
TranSSL$^*$~\cite{TranSSL}  & ViT-S & 89.6 & 95.3 & 50.6 & 75.0 \\
\rowcolor{lightgray} 
Ours & R50 & \textbf{90.6} & \textbf{95.7} & \textbf{52.5} & \textbf{78.3} \\
\rowcolor{lightgray} 
Ours & R50-IBN & \textbf{91.4} & \textbf{96.1} & \textbf{59.4} & \textbf{82.4} \\
\rowcolor{lightgray}
Ours  & ViT-S & \textbf{90.6} & \textbf{95.6} & \textbf{55.0} & \textbf{78.4} \\
\bottomrule[1pt]
\end{tabular}
}
\end{minipage}
\vspace{-8mm}
\end{table}

\textbf{Unsupervised person ReID.} We compare our CION with some SoTA methods of unsupervised person ReID in Tab.~\ref{table:unsupervised}, which contains two settings, \textit{i.e.}, unsupervised domain adaptation (UDA) and unsupervised learning (USL). All the person-centric pre-trained models are fine-tuned with C-Contrast~\cite{CContrast}. For both settings, our method achieves new SoTA performance, significantly surpassing all previous methods. For example, with ResNet50-IBN as backbone, our method outperforms the previous SoTA LUP~\cite{TranSSL} by a large margin, \textit{i.e.}, 20.1\% mAP on Market1501 $\rightarrow$ MSMT17 UDA setting and 19.6\% mAP on MSMT17 USL setting, respectively. Noting that LUP is an instance-level method, these results verify the superiority of our identity-level method in person ReID pre-training.

\vspace{-3mm}
\subsection{Generalizing to different model structures}
\begin{wraptable}[33]{r}{9cm}
\vspace{-6mm}
\tiny
\tabcolsep=3pt
\begin{center}
\caption{CION significantly improves different models. FLOPs are computed based on the input size. * means the result with the input size as $384\times128$. $^{\dag}$ represents fine-tuning with MGN.
$\uparrow$ indicates the improvement brought by our CION over the baseline.}
\label{tab:modelzoo}
\begin{threeparttable}
\begin{tabular}{lcccccc}
\toprule[1pt]
 \multirow{2}{*}{\textbf{Model}} & \multirow{1}{*}{\textbf{Params.}} &\multirow{1}{*}{\textbf{FLOPs}} & \multicolumn{2}{c}{\textbf{Market1501}}  & \multicolumn{2}{c}{\textbf{MSMT17}}  \\ \cline{4-7}
        & \textbf{(M)} &  \textbf{(G)}& \textbf{mAP} & \textbf{Rank1} & \textbf{mAP} & \textbf{Rank1} \\ 
       \hline 
  ResNet50$^{\dag*}$~\cite{ResNet} & 23.51&6.14&92.3($\uparrow$\textbf{4.7}) & 97.1($\uparrow$\textbf{2.1}) & 70.1($\uparrow$\textbf{6.6}) & 87.8($\uparrow$\textbf{2.8})  \\
  ResNet101$^{\dag*}$~\cite{ResNet} & 42.50&9.80&93.1($\uparrow$\textbf{4.9}) & 97.1($\uparrow$\textbf{1.8}) & 71.7($\uparrow$\textbf{6.9}) & 88.8($\uparrow$\textbf{2.9})  \\
  ResNet152$^{\dag*}$~\cite{ResNet} & 58.14&13.47&93.1($\uparrow$\textbf{4.4}) & 97.3($\uparrow$\textbf{1.6}) & 72.4($\uparrow$\textbf{6.5}) & 89.2($\uparrow$\textbf{3.1})  \\
  ResNet50-IBN$^{\dag*}$~\cite{IBNNet} & 23.51&6.14&93.3($\uparrow$\textbf{4.1}) & 97.3($\uparrow$\textbf{1.5}) & 74.3($\uparrow$\textbf{8.9}) & 89.8($\uparrow$\textbf{3.1})  \\
  ResNet101-IBN$^{\dag*}$~\cite{IBNNet} & 42.50&9.80&93.8($\uparrow$\textbf{4.3}) & 97.2($\uparrow$\textbf{1.3}) & 76.0($\uparrow$\textbf{8.4}) & 90.7($\uparrow$\textbf{3.1})  \\
  ResNet152-IBN$^{\dag*}$~\cite{IBNNet} & 58.14&13.47&94.3($\uparrow$\textbf{4.5}) & 97.3($\uparrow$\textbf{1.6}) & 76.8($\uparrow$\textbf{8.6}) & 90.8($\uparrow$\textbf{3.3})  \\
  \hdashline[2.5pt/5pt]
  GhostNet 0.5$\times$~\cite{GhostNet} & 1.31&0.03&81.0($\uparrow$\textbf{10.7}) & 92.0($\uparrow$\textbf{5.8}) & 39.6($\uparrow$\textbf{10.6}) & 65.4($\uparrow$\textbf{9.1})  \\
  GhostNet 1.0$\times$~\cite{GhostNet}& 3.90&0.10&87.7($\uparrow$\textbf{11.5}) & 95.1($\uparrow$\textbf{6.1}) & 53.7($\uparrow$\textbf{12.0}) & 76.9($\uparrow$\textbf{9.8})  \\
  GhostNet 1.3$\times$~\cite{GhostNet}& 6.08&0.16&88.8($\uparrow$\textbf{12.4})& 95.2($\uparrow$\textbf{5.9}) & 55.4($\uparrow$\textbf{11.7}) & 78.1($\uparrow$\textbf{9.4})  \\
  EdgeNext-XS~\cite{EdgeNext} & 2.14&0.27&88.7($\uparrow$\textbf{15.9}) & 94.7($\uparrow$\textbf{6.5}) & 57.9($\uparrow$\textbf{18.1}) & 79.8($\uparrow$\textbf{13.7})  \\
  EdgeNext-S~\cite{EdgeNext} & 5.28&0.63&91.3($\uparrow$\textbf{14.5}) & 96.1($\uparrow$\textbf{5.9}) & 63.2($\uparrow$\textbf{19.8}) & 83.2($\uparrow$\textbf{13.7})  \\
  EdgeNext-B~\cite{EdgeNext} & 17.93&1.92&92.4($\uparrow$\textbf{14.7}) & 96.8($\uparrow$\textbf{6.5}) & 68.6($\uparrow$\textbf{22.3}) & 86.6($\uparrow$\textbf{14.7})  \\
  RepViT-M0.9~\cite{RepViT} & 4.72&0.56&92.8($\uparrow$\textbf{12.8}) & 96.9($\uparrow$\textbf{5.8}) & 68.4($\uparrow$\textbf{19.8}) & 86.5($\uparrow$\textbf{12.0})  \\
  RepViT-M1.0~\cite{RepViT} & 6.40&0.75&93.1($\uparrow$\textbf{25.6}) & 96.6($\uparrow$\textbf{11.4}) & 69.5($\uparrow$\textbf{47.5}) & 87.2($\uparrow$\textbf{43.4})  \\
  RepViT-M1.5~\cite{RepViT} & 13.62&1.54&92.8($\uparrow$\textbf{23.2}) & 96.6($\uparrow$\textbf{10.3}) & 69.5($\uparrow$\textbf{35.1}) & 87.5($\uparrow$\textbf{26.2})  \\
  FastViT-S12~\cite{FastViT} & 8.45&0.93&91.2($\uparrow$\textbf{12.3}) & 96.4($\uparrow$\textbf{5.0}) & 59.0($\uparrow$\textbf{17.3}) & 81.4($\uparrow$\textbf{13.6})  \\
  FastViT-SA12~\cite{FastViT} & 10.56&1.00&90.9($\uparrow$\textbf{14.7}) & 95.8($\uparrow$\textbf{5.9}) & 59.0($\uparrow$\textbf{17.1}) & 81.6($\uparrow$\textbf{13.8})  \\
  FastViT-SA24~\cite{FastViT} & 20.53&1.92&92.9($\uparrow$\textbf{19.9}) & 96.9($\uparrow$\textbf{8.7}) & 69.6($\uparrow$\textbf{32.6}) & 87.6($\uparrow$\textbf{23.8})  \\
  ResNet18~\cite{ResNet} & 11.18&2.00&89.3($\uparrow$\textbf{10.9}) & 95.6($\uparrow$\textbf{4.5}) & 55.6($\uparrow$\textbf{14.8}) & 79.6($\uparrow$\textbf{12.0})  \\
  ResNet50~\cite{ResNet} & 23.51&4.09&92.8($\uparrow$\textbf{9.9}) & 96.8($\uparrow$\textbf{3.9}) & 67.7($\uparrow$\textbf{15.5}) & 86.4($\uparrow$\textbf{9.8})  \\
  ResNet101~\cite{ResNet} & 42.50&6.54&94.0($\uparrow$\textbf{10.8}) & 97.3($\uparrow$\textbf{4.6}) & 73.2($\uparrow$\textbf{17.3}) & 88.8($\uparrow$\textbf{9.0})  \\
  ResNet152~\cite{ResNet} & 58.14&8.98&94.2($\uparrow$\textbf{8.1}) & 97.1($\uparrow$\textbf{3.0}) & 74.5($\uparrow$\textbf{18.2}) & 89.3($\uparrow$\textbf{10.4})  \\
  ResNet18-IBN~\cite{IBNNet} & 11.18&2.00&90.0($\uparrow$\textbf{10.2}) & 96.0($\uparrow$\textbf{4.4}) & 56.1($\uparrow$\textbf{13.5}) & 79.5($\uparrow$\textbf{10.4})  \\
  ResNet50-IBN~\cite{IBNNet} & 23.51&4.09&93.7($\uparrow$\textbf{8.9}) & 96.9($\uparrow$\textbf{3.1}) & 72.3($\uparrow$\textbf{18.6}) & 88.5($\uparrow$\textbf{11.0})  \\
  ResNet101-IBN~\cite{IBNNet} & 42.50&6.54&94.3($\uparrow$\textbf{7.9}) & 97.3($\uparrow$\textbf{3.1}) & 75.8($\uparrow$\textbf{18.0}) & 89.8($\uparrow$\textbf{9.6})  \\
  ResNet152-IBN~\cite{IBNNet} & 58.14&8.98&94.6($\uparrow$\textbf{7.2}) & 97.6($\uparrow$\textbf{3.0}) & 76.7($\uparrow$\textbf{17.8}) & 90.3($\uparrow$\textbf{8.9})  \\
  ConvNext-T~\cite{ConvNext} & 27.82&2.92&92.0($\uparrow$\textbf{14.1}) & 96.9($\uparrow$\textbf{6.0}) & 68.9($\uparrow$\textbf{22.7}) & 87.3($\uparrow$\textbf{14.8})  \\
  ConvNext-S~\cite{ConvNext} & 49.45&5.68&92.9($\uparrow$\textbf{12.5}) & 96.8($\uparrow$\textbf{5.2}) & 69.9($\uparrow$\textbf{21.8}) & 88.0($\uparrow$\textbf{14.4})  \\
  ConvNext-B~\cite{ConvNext} & 87.57&10.04&93.4($\uparrow$\textbf{13.2}) &96.9($\uparrow$\textbf{5.3}) & 72.7($\uparrow$\textbf{24.1}) & 89.0($\uparrow$\textbf{14.5})  \\
  VOLO-D1~\cite{VOLO} & 25.84&4.40&93.0($\uparrow$\textbf{7.8}) & 96.9($\uparrow$\textbf{3.3}) & 73.9($\uparrow$\textbf{15.8}) & 88.7($\uparrow$\textbf{8.6})  \\
  VOLO-D2~\cite{VOLO} & 57.62&9.15&92.1($\uparrow$\textbf{7.1}) & 96.4($\uparrow$\textbf{2.9}) & 70.9($\uparrow$\textbf{11.6}) & 87.5($\uparrow$\textbf{6.6})  \\
  VOLO-D3~\cite{VOLO} & 85.26&13.26&90.5($\uparrow$\textbf{5.6}) & 95.7($\uparrow$\textbf{1.8}) & 62.8($\uparrow$\textbf{2.1}) & 83.0($\uparrow$\textbf{1.3})  \\
  ViT-T~\cite{ViT} & 6.43&1.55&91.4($\uparrow$\textbf{10.5}) & 96.3($\uparrow$\textbf{5.1}) & 65.0($\uparrow$\textbf{18.9}) & 84.6($\uparrow$\textbf{14.9})  \\
  ViT-S~\cite{ViT} & 23.39&3.79&92.8($\uparrow$\textbf{8.0}) & 96.5($\uparrow$\textbf{2.6}) & 70.3($\uparrow$\textbf{16.7}) & 86.9($\uparrow$\textbf{11.6})  \\
  ViT-B~\cite{ViT} & 89.15&12.37&93.1($\uparrow$\textbf{6.4}) & 96.9($\uparrow$\textbf{2.3}) & 72.6($\uparrow$\textbf{11.4}) & 88.5($\uparrow$\textbf{6.6})  \\
  Swin-T~\cite{Swin} & 27.53&3.13&92.5($\uparrow$\textbf{9.8}) & 97.0($\uparrow$\textbf{4.3}) & 71.1($\uparrow$\textbf{19.6}) & 87.6($\uparrow$\textbf{12.6})  \\
  Swin-S~\cite{Swin} & 48.85&5.93&93.6($\uparrow$\textbf{9.1}) & 96.8($\uparrow$\textbf{3.6}) & 76.0($\uparrow$\textbf{19.7}) & 89.6($\uparrow$\textbf{10.9})  \\
  Swin-B~\cite{Swin} & 86.76&10.44&94.0($\uparrow$\textbf{9.6}) & 96.9($\uparrow$\textbf{3.0}) & 76.7($\uparrow$\textbf{20.7}) & 90.3($\uparrow$\textbf{12.1})  \\
  \hline
  Average Impr. & &&$\uparrow$\textbf{11.7} & $\uparrow$\textbf{5.0} & $\uparrow$\textbf{18.8} & $\uparrow$\textbf{12.9}  \\
\Xhline{1pt}
\end{tabular}
\end{threeparttable}
\end{center}
\end{wraptable}
\vspace{-1mm}
During pre-training, CION imposes minimal constraints on the model structure, allowing for its generalization across a variety of structures. To verify this, we conduct extensive experiments to investigate whether CION can improve the models with different structures and parameters, encompassing 32 models with 10 structures, including CNNs, ViTs, and other specialized architectures. The baselines are the models with supervised ImageNet~\cite{ImageNet} pre-training. As a common practice, we utilize TransReID~\cite{TransReID} to fine-tune the models on ReID datasets with input size as $256\times128$, if not specified. As the results listed in Tab.~\ref{tab:modelzoo}, our CION consistently yields significant performance improvements across all models. When applying to TransReID, the average performance improvements of all the 32 models over the baselines are 11.7\% and 18.8\% mAP on Market1501 and MSMT17, respectively. With CION demonstrating such model-agnostic ability, we construct a model zoo named ReIDZoo, which contains all the CION pre-trained models with spanning structures and parameters, to meet diverse research and application needs in this field.

\vspace{-5mm}
\subsection{Ablation studies and analyses}
\vspace{-1mm}
\begin{wrapfigure}[19]{r}{7cm}
\vspace{-4mm}
\centering
\includegraphics[width=\linewidth]{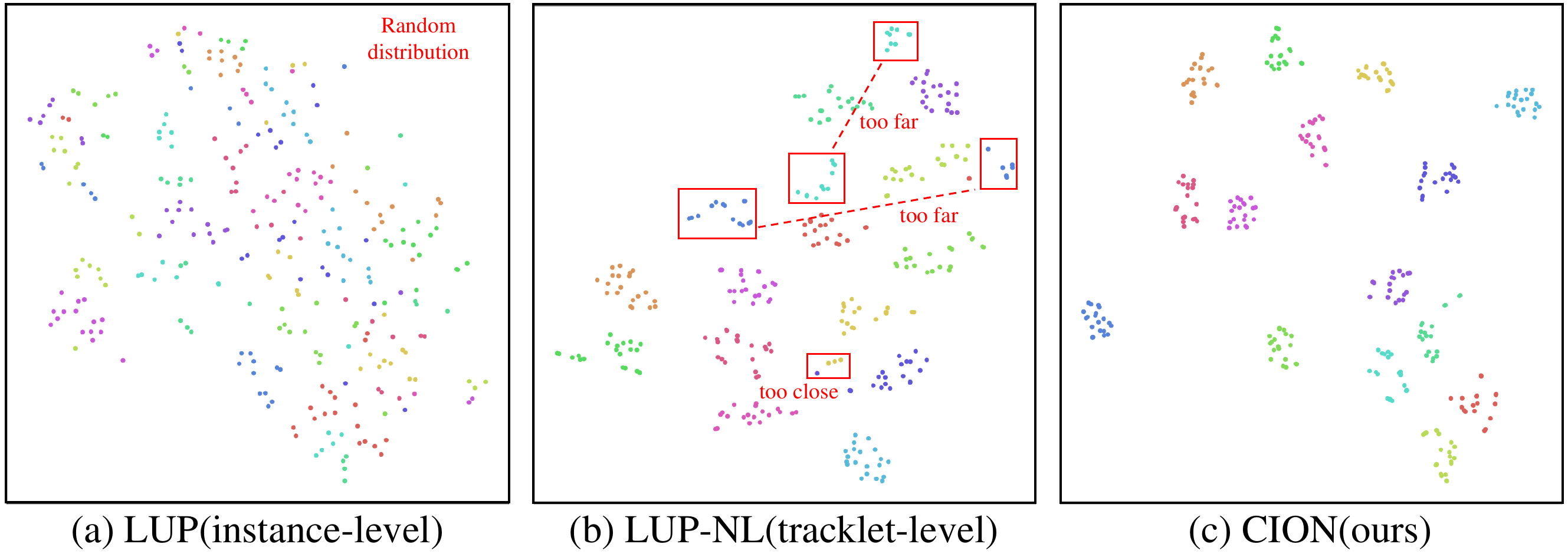}
\caption{The t-SNE~\cite{tSNE} visualization of extracted image features. Our CION enjoys good identity consistency and discrimination, while the instance-level method LUP~\cite{LUP} and tracklet-level method LUP-NL~\cite{LUPNL} do not (marked in red).}
\label{fig:tSNE}
\end{wrapfigure}
\textbf{Identity-level correlating holds significance.} We validate the significance of identity-level correlating in learning better person representations through both feature distribution visualization and performance comparison. For visualization, we collect samples of 15 persons randomly selected from the gallery of Market1501~\cite{Market1501}, and visualize their features via t-SNE~\cite{tSNE} in Fig.~\ref{fig:tSNE}. The instance-level method LUP~\cite{LUP} and the tracklet-level method LUP-NL~\cite{LUPNL}, which are respectively pre-trained on 4.2M and 10.7M person images, are employed as a comparison. As we can see, the features of LUP are almost randomly distributed, while the features of LUP-NL remain challenging to aggregate at the identity level. However, our CION consistently enjoys good intra-consistency and inter-discrimination.
\begin{wraptable}[5]{r}{7cm}
\vspace{-23mm}
\small
\tabcolsep=2pt
\begin{center}
\caption{Our identity-level method achieves significant leading performance compared with two identity-ignored pre-training methods~\cite{LUP,DINO}.}
\vspace{-1mm}
\label{tab:ablation_idino}
\begin{threeparttable}
\begin{tabular}{lccccc}
\toprule[1pt]
\multirow{2}{*}{\textbf{Method}} & \multirow{2}{*}{\textbf{Pre-train}}& \multicolumn{2}{c}{\textbf{Market1501}}  & \multicolumn{2}{c}{\textbf{MSMT17}}  \\ \cline{3-6}
       &  & \textbf{mAP} & \textbf{Rank1} & \textbf{mAP} & \textbf{Rank1} \\ 
       \hline 
LUP+MGN &CION-AL&91.4 & 96.6 & 68.2 & 86.4 \\

DINO+MGN   & CION-AL&90.9 & 96.3 & 66.7 & 85.8 \\
\rowcolor{lightgray}
Ours+MGN  & CION-AL&\textbf{93.3} & \textbf{97.3} & \textbf{74.3} & \textbf{89.8} \\
\bottomrule[1pt]
\end{tabular}
\end{threeparttable}
\end{center}
\end{wraptable}

\vspace{-1.5mm}
For performance comparison, we compare the fine-tuning results of our CION with two popular identity-ignored pre-training methods LUP~\cite{LUP} and DINO~\cite{DINO} in Tab.~\ref{tab:ablation_idino}. For a fair comparison, the models are all ResNet50-IBN~\cite{IBNNet} pre-trained on our CION-AL and the fine-tuning algorithm is MGN~\cite{MGN}. We can see that the model pre-trained with CION achieves significant leading performance. These results demonstrate the significance of implementing identity-level correlating for person ReID pre-training.

\begin{wrapfigure}[12]{r}{7cm}
\vspace{-4.5mm}
\centering
\includegraphics[width=\linewidth]{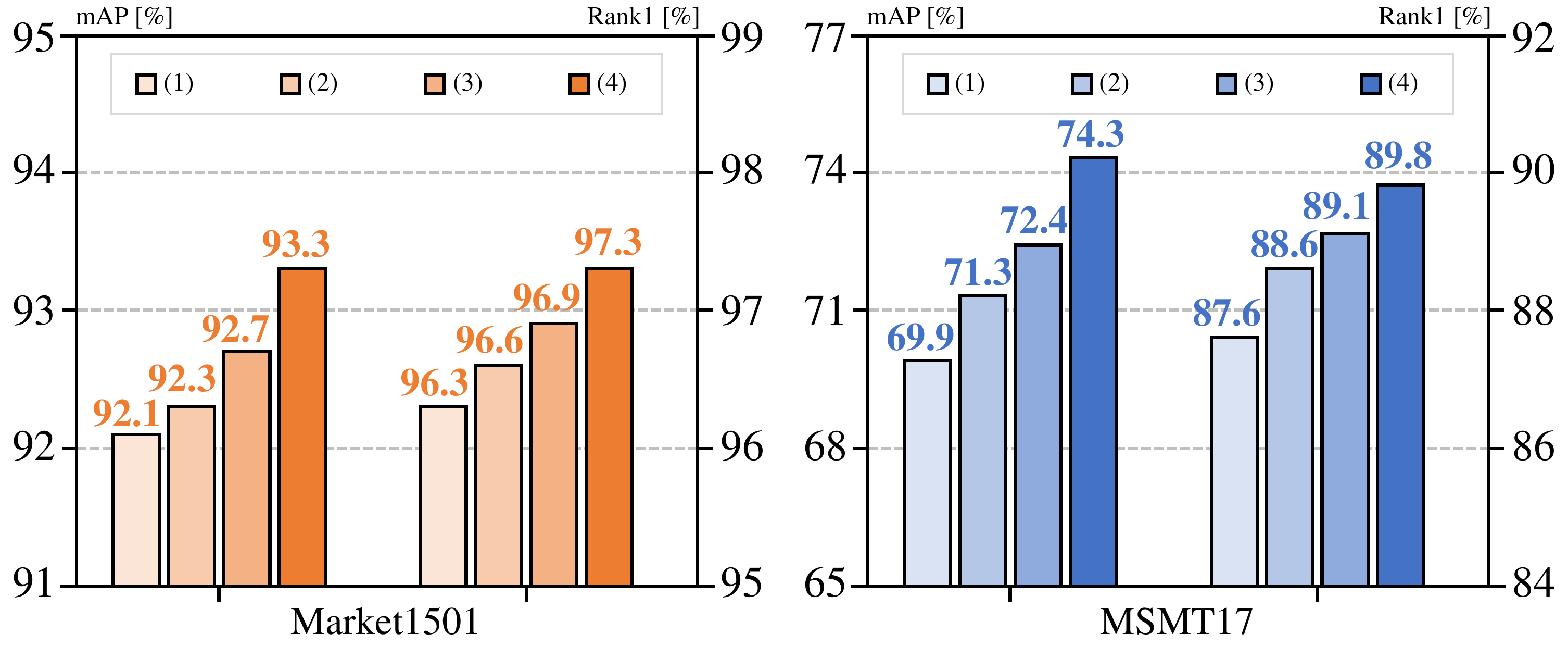}
\vspace{-6mm}
\caption{Results with different denoising strategies: (1) without denoising, (2) add single-tracklet denoising, (3) further add single-video denoising, (4) further add cross-video denoising. }
\label{fig:denoising}
\end{wrapfigure}
\vspace{-1mm}
\textbf{Effectiveness of progressive multi-level denoising.}
Fig.~\ref{fig:denoising} shows the ablation study of each denoising strategy. The number of training samples for all groups is set to 3.9M for a fair comparison. The backbone is ResNet50-IBN~\cite{IBNNet} and the fine-tuning algorithm is MGN~\cite{MGN}. We can observe that each strategy significantly improves the fine-tuning performance of the pre-trained model on downstream datasets. Particularly, the cross-video denoising results in the most substantial improvement, further increasing the mAP by 0.6\% and 1.9\% on Market1501 and MSMT17, respectively. These results verify the effectiveness of our progressive multi-level denoising strategy, which guarantees satisfactory identity correlation seeking.

\begin{wrapfigure}[9]{r}{7cm}
\vspace{-4mm}
\centering
\includegraphics[width=\linewidth]{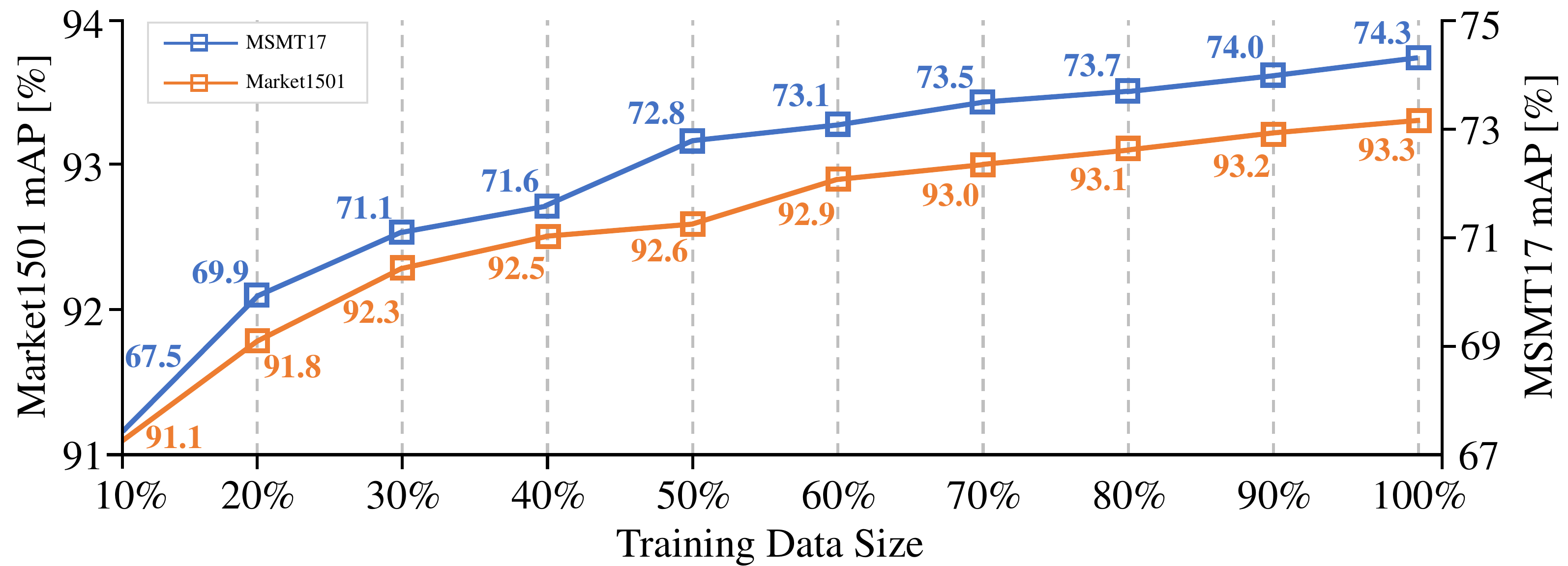}
\vspace{-6mm}
\caption{The fine-tuning performance consistently keeps improving as the training data increases.  }
\label{fig:scalability}
\end{wrapfigure}
\textbf{Scalability to large-scale training data.} 
We also study the scalability of our method to large-scale training data. Aside from the training data size, the rest settings are same with the experiment mentioned above. As shown in Fig~\ref{fig:scalability}, with training data size ratio varying from 10\% to 100\%, the fine-tuning performance on downstream datasets consistently keeps improving, especially on the more challenging MSMT17. This indicates our method’s potential for further improved performance with increasing training data.

\vspace{-4mm}
\section{Conclusion}
\vspace{-3mm}
We present a novel framework, CION, to learn identity-invariance from cross-video person images for person re-identification pre-training. In particular, we model the identity correlation seeking process as a progressive multi-level denoising problem, with a novel noise concept defined. Also, we propose an identity-guided self-distillation loss to implement better large-scale pre-training. Compared to existing pre-training methods, we achieve significantly leading performance with even fewer training samples. Meanwhile , we contribute a model zoo to promote the development of this field.

\noindent
\textbf{Acknowledgements.} This work was supported by the National Natural Science Foundation of China No.62176097, and the Hubei Provincial Natural Science Foundation of China No.2022CFA055. We gratefully acknowledge the support of MindSpore, CANN (Compute Architecture for Neural Networks) and Ascend AI Processor used for this research.
\bibliographystyle{plain}
\bibliography{ref}

\newpage
\appendix

\section{Appendix}
\subsection{Broader impact}
\label{broadimpact}
This paper proposes a cross-video identity-correlating pre-training framework for person re-identification, and contributes a pre-trained model zoo with spanning model structures and parameters. Our framework and models can help to identity and track pedestrians across different cameras, thus boosting the development of smart retail, smart transportation, smart security systems and so on in the future metropolises. In addition, our proposed pre-training framework is quite general and not limited to the specific research field of person ReID. It can be well extended to broader research areas with identity concept such as vehicle ReID.

Nevertheless, the application of person ReID, such as for identifying and tracking pedestrians in surveillance systems~\cite{gong2022person,gong2024cross,shi2024learningcommonalitydivergencevariety,dpis,mmm}, might raise privacy concerns. It typically depends on the utilization of surveillance data for training without the explicit consent of the individuals being recorded. Therefore, governments and officials need to carefully establish strict regulations and laws to control the usage of ReID technologies. Otherwise, person ReID technologies~\cite{RPLNR,AIO,li2021weperson,ye2021collaborative} can potentially equip malicious actors with the ability to surveil pedestrians through multiple cameras without their consent. Meanwhile, the research community should also avoid using any dataset with ethics issues. For example, DukeMTMC~\cite{DukeMTMC} has been taken down due to the violation of data collection terms and should no longer be used. We would not evaluate our models on DukeMTMC related benchmarks as well. Furthermore, we should be cautious of the misidentification of the ReID systems to avoid possible disturbance. Also, note that the demographic makeup of the datasets used is not representative of the broader population.

At the same time, we have utilized a substantial amount of person-containing video data from the internet for pre-training purposes. Consequently, it is inevitable that the resulting models may inherently contain information about these persons. Researchers should adhere to relevant laws and regulations, and strive to avoid using our models for any improper invasion of privacy. We will have a gated release of our models and training data to avoid any misuse. We will require that users adhere to usage guidelines and restrictions to access our models and training data. Meanwhile, all our open-sourced assets can only be used for research purpose and are forbidden for any commercial use.

\subsection{Limitations}
\label{limitations}
CION presents a preliminary attempt to correlate person images across different internet videos to implement person re-identification pre-training. 
Despite its effectiveness on existing public datasets, CION may still be difficult to learn good fine-grained person representations for it does not explicitly achieve further fine-grained information mining. 
Meanwhile, due to the limitations of the feature extractor’s representational capacity, it is inevitable that some extremely difficult noise will be hard to remove during the denoising process. As a result, the automatically sought identity correlation may still have a certain discrepancy in accuracy when compared to manually annotated correlation, which may introduce a certain degree of ambiguity during the training process. Additionally, the Sliding Range and Linking Relation strategy we proposed to conserve computational resources may result in difficulty seeking correlations between tracklets of the same person across different videos that are spaced extremely far apart.
Also, as we have followed the conventional practice of previous works by utilizing a large amount of person-containing internet video data to implement pre-training, there is a potential for privacy and security issues to some extent. 
Therefore, in our subsequent work, we will focus on addressing fine-grained issues and improving the accuracy of identity correlations. We will take every possible measure to prevent the misuse of our models and dataset as well.

\subsection{Safeguards for dataset and models}
\label{safeguards}
To address the privacy concerns associated with crawling videos from the internet and the application of person ReID technology, we will implement a controlled release of our models and dataset, thereby preventing privacy violations and ensuring information security. We require that the crawled images are sourced from legitimate video websites to avoid the inclusion of harmful content. We will also require that users adhere to usage guidelines and restrictions to access our models and dataset. For instance, we have drafted the following regulations that must be adhered to, which will be refined and elaborated in subsequent releases: 

1. Privacy: All individuals using the models in ReIDZoo and the CION-AL dataset should agree to protect the privacy of all the subjects in it. The users should bear all responsibilities and consequences for any loss caused by the misuse. 

2. Redistribution: The models in ReIDZoo and the CION-AL dataset, either entirely or partly, should not be further distributed, published, copied, or disseminated in any way or form without a prior approval from the creators, no matter for profitable use or not.

3. Commercial Use: The models in ReIDZoo and the CION-AL dataset, entirely, partly, or in any format derived thereof, is not allowed for commercial use.

4. Modification: The models in ReIDZoo and the CION-AL dataset, either entirely or partly, is not allowed to be modified.

In parallel, we will require users to provide relevant information and will rigorously screen the submitted details to restrict access to our dataset and models by institutions or individuals with a history of privacy violations.

\vspace{-5mm}
\subsection{Licenses for existing assets}
\label{license}
\vspace{-2mm}
We utilize some existing assets to conduct our research, which can be categorized to three types: code, data and models. All assets used in our paper are properly credited and all original paper are cited. We list the license and URL for each asset below.

\textbf{Code}

1. Emerging Properties in Self-Supervised Vision Transformers~\cite{DINO}. (DINO)

Apache 2.0 License, URL:~\url{https://github.com/facebookresearch/dino}

2. TransReID: Transformer-based Object Re-Identification~\cite{TransReID}. (TransReID)

MIT License, URL:~\url{https://github.com/damo-cv/TransReID}

3. Cluster Contrast for Unsupervised Person Re-Identification~\cite{CContrast}. (C-Contrast)

MIT License, URL:~\url{https://github.com/alibaba/cluster-contrast-reid}

4. Unsupervised Pre-training for Person Re-identification~\cite{LUP}. (LUP)

MIT License, URL:~\url{https://github.com/DengpanFu/LUPerson}

5. Self-Supervised Pre-Training for Transformer-Based Person Re-Identification~\cite{TranSSL}. (TranSSL)

MIT License, URL:~\url{https://github.com/damo-cv/TransReID-SSL}

6. FastReID: A Pytorch Toolbox for General Instance Re-identification~\cite{fastreid}. (FastReID)

Apache 2.0 License, URL:~\url{https://github.com/JDAI-CV/fast-reid}

\textbf{Data}

1. PLIP: Language-Image Pre-training for Person Representation Learning~\cite{PLIP}. (SYNTH-PEDES)

MIT License, URL:~\url{https://github.com/Zplusdragon/PLIP}

2. UFineBench: Towards Text-based Person Retrieval with Ultra-fine Granularity~\cite{UFineBench}. (UFine6926)

UFineBench License, URL:~\url{https://github.com/Zplusdragon/UFineBench}

3. Person Transfer GAN to Bridge Domain Gap for Person Re-Identification~\cite{MSMT17}. (MSMT17)

MSMT17 Release Agreement, URL:~\url{https://www.pkuvmc.com/dataset.html}

4. Scalable Person Re-Identification: A Benchmark~\cite{Market1501}. (Market1501)

MIT License, URL:~\url{http://zheng-lab.cecs.anu.edu.au/Project/project_reid.html}

\textbf{Models}

1. Deep Residual Learning for Image Recognition~\cite{ResNet}. (ResNet)

BSD-3-Clause license, URL:~\url{https://github.com/pytorch/vision}

2. Instance-Batch Normalization Network~\cite{IBNNet}. (ResNet-IBN)

MIT License, URL:~\url{https://github.com/XingangPan/IBN-Net}

3.  GhostNet: More Features from Cheap Operations~\cite{GhostNet}. (GhostNet)

Apache 2.0 License, URL:~\url{https://github.com/huawei-noah/Efficient-AI-Backbones/}

4. EdgeNeXt: Efficiently Amalgamated CNN-Transformer Architecture for Mobile Vision Applications~\cite{EdgeNext}. (EdgeNext)

MIT License: URL:~\url{https://github.com/mmaaz60/EdgeNeXt}

5. RepViT: Revisiting Mobile CNN From ViT Perspective~\cite{RepViT}. (RepViT)

Apache 2.0 License, URL:~\url{https://github.com/THU-MIG/RepViT}

6. FastViT: A Fast Hybrid Vision Transformer using Structural Reparameterization~\cite{FastViT}. (FastViT)

FastViT License, URL:~\url{https://github.com/apple/ml-fastvit}

7. A ConvNet for the 2020s~\cite{ConvNext}. (ConvNext)

MIT License, URL:~\url{https://github.com/facebookresearch/ConvNeXt}

8. An Image is Worth 16x16 Words: Transformers for Image Recognition at Scale. (ViT)

Apache 2.0 License, URL:~\url{https://github.com/google-research/vision_transformer}

9. Swin Transformer: Hierarchical Vision Transformer using Shifted Windows~\cite{Swin}. (Swin)

MIT License, URL:~\url{https://github.com/microsoft/Swin-Transformer}

10. VOLO: Vision Outlooker for Visual Recognition~\cite{VOLO}. (VOLO)

Apache 2.0 License, URL:~\url{https://github.com/sail-sg/volo}

\vspace{-2mm}
\subsection{Pre-training setting for each model}
\label{pretrainingsetting}
\begin{wraptable}[26]{r}{7cm}
\small
\vspace{-17mm}
\tabcolsep=2pt
\begin{center}
\caption{The batch sizes, numbers of local cropped views and rough training time for each model.}
\vspace{-3mm}
\label{tab:eachmodeldetail}
\begin{threeparttable}
\begin{tabular}{lccc}
\toprule[1pt]
Model&batch size&local views &time(d$\approx$)\\
       \hline 
 GhostNet 0.5$\times$~\cite{GhostNet} & 216&8&4.8  \\
  GhostNet 1.0$\times$~\cite{GhostNet}& 192&8&5.7  \\
  GhostNet 1.3$\times$~\cite{GhostNet}& 168&8&7.2  \\
  EdgeNext-XS~\cite{EdgeNext} & 192&8&5.1 \\
  EdgeNext-S~\cite{EdgeNext} & 144&8&6.5  \\
  EdgeNext-B~\cite{EdgeNext} & 120&8&7.3  \\
  RepViT-M0.9~\cite{RepViT} & 144&8&5.1 \\
  RepViT-M1.0~\cite{RepViT} & 144&8&6.2  \\
  RepViT-M1.5~\cite{RepViT} & 96&8&8.1 \\
  FastViT-S12~\cite{FastViT} &120&8&5.3  \\
  FastViT-SA12~\cite{FastViT} & 120&8&6.8   \\
  FastViT-SA24~\cite{FastViT} & 96&8&9.3  \\
  ResNet18~\cite{ResNet} & 192&10&3.6 \\
  ResNet50~\cite{ResNet} & 120&8&5.1 \\
  ResNet101~\cite{ResNet} & 96&8&6.6  \\
  ResNet152~\cite{ResNet} &72&8&9.7 \\
  ResNet18-IBN~\cite{IBNNet} & 192&10&3.6 \\
  ResNet50-IBN~\cite{IBNNet} & 120&8&5.1  \\
  ResNet101-IBN~\cite{IBNNet} & 96&8&6.6\\
  ResNet152-IBN~\cite{IBNNet} & 72&8&9.7\\
  ConvNext-T~\cite{ConvNext} & 120&7&6.2 \\
  ConvNext-S~\cite{ConvNext} & 96&6&7.3 \\
  ConvNext-B~\cite{ConvNext} & 72&6&9.5\\
  ViT-T~\cite{ViT} & 120&8&5.3 \\
  ViT-S~\cite{ViT} & 96&8&6.7 \\
  ViT-B~\cite{ViT} & 72&8&8.6  \\
  Swin-T~\cite{Swin} &96&7&5.8  \\
  Swin-S~\cite{Swin} & 72&6&8.3  \\
  Swin-B~\cite{Swin} & 48&8&11.6 \\
  VOLO-D1~\cite{VOLO} & 120&4&6.2   \\
  VOLO-D2~\cite{VOLO} & 72&4&9.4  \\
  VOLO-D3~\cite{VOLO} & 48&4&13.3   \\
\bottomrule[1pt]
\end{tabular}
\end{threeparttable}
\end{center}
\end{wraptable}
Due to the varying memory requirements of different models during the pre-training process, we optimize the utilization of computational resources by setting different batch sizes and numbers of local cropped views for each model. We report the batch sizes, numbers of local cropped views and rough training time for each model in Tab.~\ref{tab:eachmodeldetail}. Each model is trained on 8 $\times$ V100 GPUs. The pre-training dataset is the whole CION-AL, which has 3,898,086 images of 246,904 persons.

\subsection{Samples used in t-SNE visualization}
\begin{figure}[h]
\vspace{-1mm}
\centering
\includegraphics[width=0.62\linewidth]{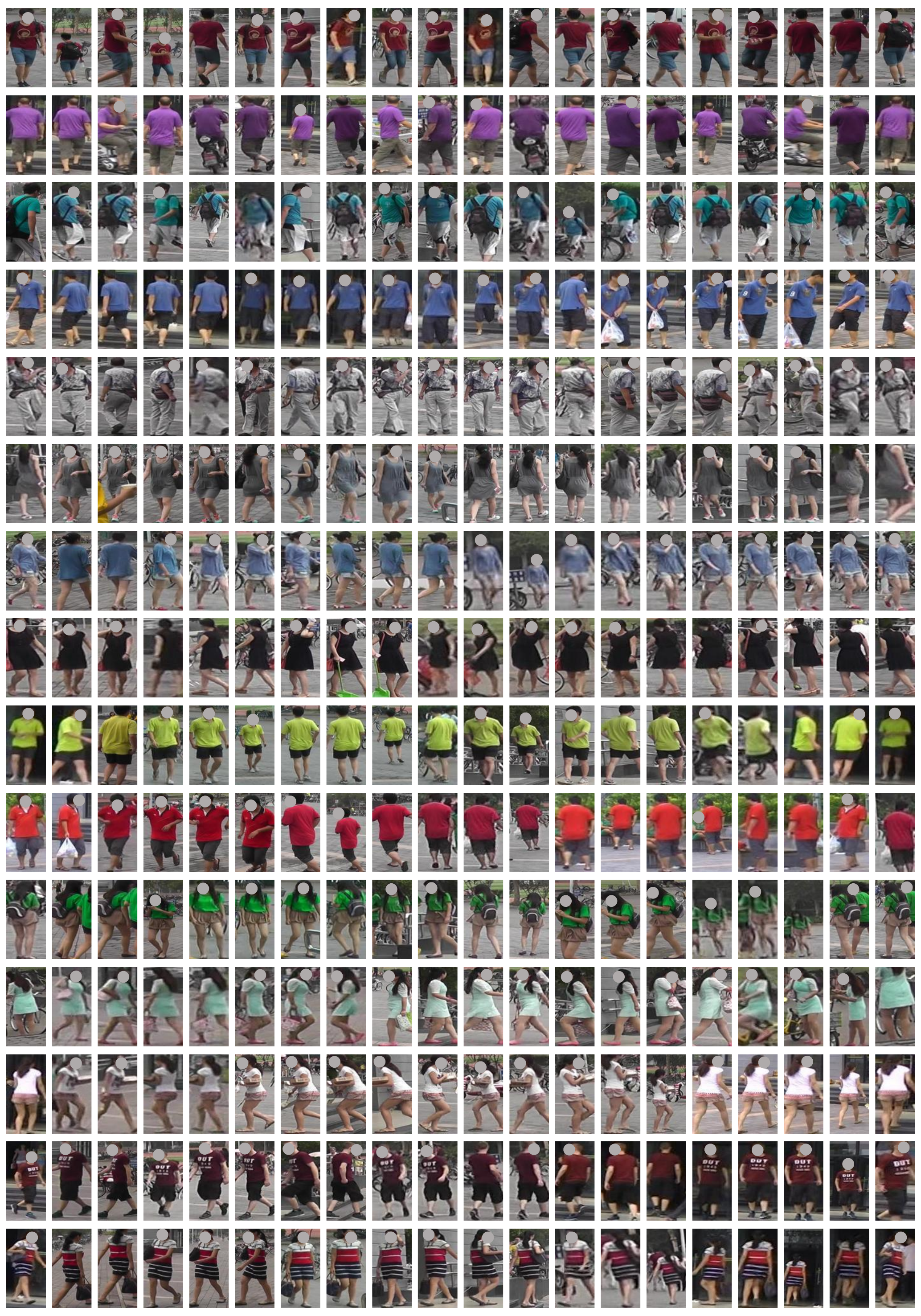}
    \caption{The samples used in t-SNE feature distribution visualization.}
    \vspace{-6mm}
\label{tSNE_samples}
\end{figure}

For reproducing the t-SNE visualization results shown in Fig.~\ref{fig:tSNE}, we display the samples we randomly selected from Market1501~\cite{Market1501} in Fig.~\ref{tSNE_samples}. There are totally 15 persons with 20 cropped images each. We obscure the faces of each person to avoid the privacy violations. Researchers can use these samples and the official models of LUP~\cite{LUP} and LUP-NL~\cite{LUPNL} to reproduce the visualization results.

\subsection{Linear training time increasing}
We have assessed how the training time of CION scale with pre-training dataset size. We use ResNet50 to conduct the experiment. We record the training time required by CION as the pre-training data size increases from 10\% to 100\%. 

\begin{wrapfigure}[7]{r}{7cm}
\vspace{-11mm}
\centering
\includegraphics[width=\linewidth]{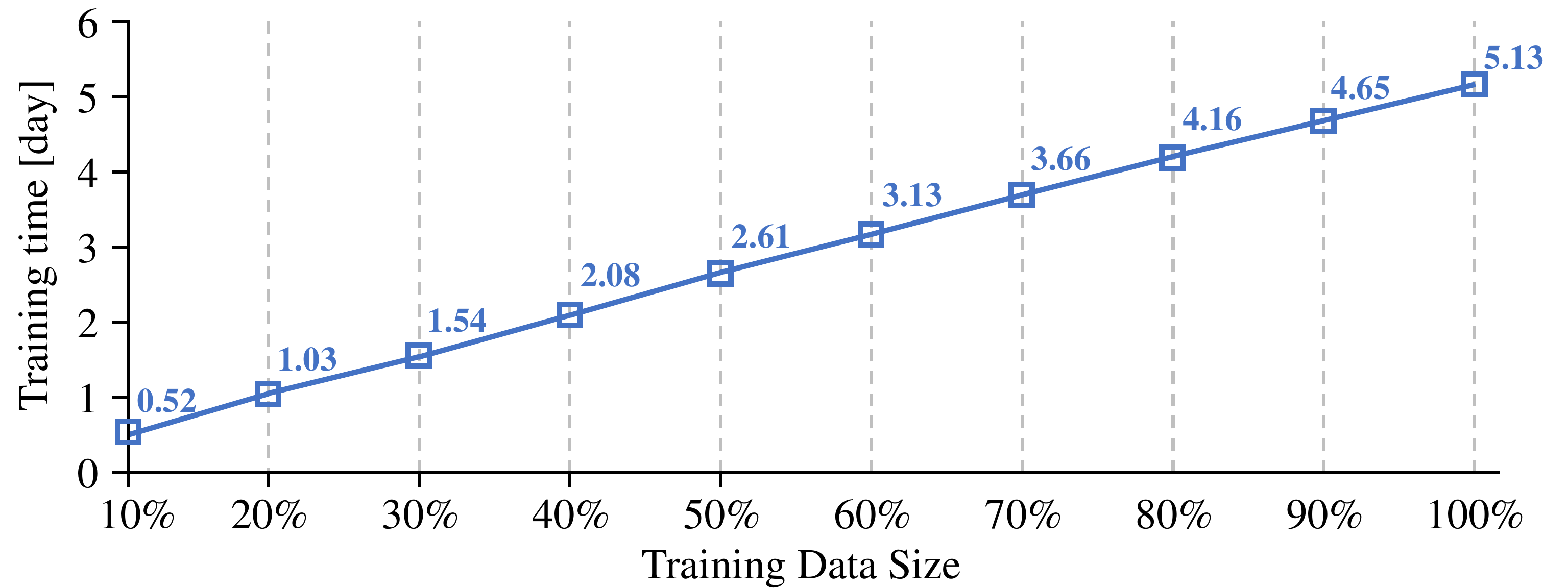}
\caption{The time consumption for pre-training on CION-AL with different data size ratio.}
\label{fig:training}
\end{wrapfigure}
\vspace{-2mm}
As shown in Fig.~\ref{fig:training}, the training time of our CION increases almost linearly with the growth in the size of the pre-trained dataset. This linear increase in training time ensures that our CION can be effectively applied to large-scale pre-training. This indicates that we can achieve better performance by adding more training data without incurring an unbearable training cost.

\vspace{-3mm}
\subsection{More results for unsupervised person ReID}
\vspace{-2mm}
\begin{wraptable}[12]{r}{8cm}
\vspace{-6mm}
\tiny
\tabcolsep=3.2pt
\begin{center}
\caption{The performance of different CION pre-trained models for unsupervised person ReID. ``Mar'' and ``MS'' denote Market1501 and MSMT17 respectively.}
\vspace{-2mm}
\label{tab:complete_unsupervised}
\begin{threeparttable}
\begin{tabular}{lcccccccc}
\toprule[1pt]
 \multirow{2}{*}{\textbf{Model}} & \multicolumn{2}{c}{\textbf{MS$\rightarrow$Mar}} &\multicolumn{2}{c}{\textbf{Mar$\rightarrow$MS}} & \multicolumn{2}{c}{\textbf{Market1501}}  & \multicolumn{2}{c}{\textbf{MSMT17}}  \\ \cline{2-9}
        & \textbf{mAP} &  \textbf{Rank1}& \textbf{mAP} & \textbf{Rank1} & \textbf{mAP} & \textbf{Rank1} & \textbf{mAP} &  \textbf{Rank1}\\ 
       \hline 
  ResNet50~\cite{ResNet} & 90.6&96.1&57.4 & 81.3 & 90.6 & 95.7 & 52.5 & 78.3   \\
  ResNet101~\cite{ResNet} & 93.1&96.8&67.6 & 86.9 & 93.2 & 97.0 & 64.2 & 85.2   \\
  ResNet152~\cite{ResNet} & 93.6&97.0&70.1 & 88.1 & 93.3 & 97.1 & 68.8 & 87.6   \\
  ResNet50-IBN~\cite{IBNNet} & 92.1&96.5&62.7 & 84.3 & 91.4 & 96.1 & 59.4 & 82.4   \\
  ResNet101-IBN~\cite{IBNNet}& 93.7&96.8&70.9 & 88.2 & 93.8 & 96.7 & 69.7 & 88.2   \\
  ResNet152-IBN~\cite{IBNNet} & 94.0&97.1&73.7 & 89.2 & 94.3 & 97.1 & 71.7 & 89.0   \\
  ViT-T~\cite{ViT} & 88.0&94.7&44.3 & 69.3 & 88.3 & 94.5 & 43.1 & 69.6   \\
  ViT-S~\cite{ViT} & 90.4&95.6&58.0 & 79.9 & 90.6 & 95.6 & 55.0 & 78.4   \\
  ViT-B~\cite{ViT} & 91.1&95.9&61.4 & 82.9 & 91.2 & 96.1 & 58.9 & 82.3   \\
\Xhline{1pt}
\end{tabular}
\end{threeparttable}
\end{center}
\end{wraptable}
To further assess the model structure adaptability of our CION, we evaluate the performance of more CION pre-trained models for unsupervised person ReID, inclucing ResNet~\cite{ResNet}, ResNet-IBN~\cite{IBNNet} and ViT~\cite{ViT}. We report the results in Tab.~\ref{tab:complete_unsupervised}. As we can see, all the models achieve superior performance for different unsupervised settings on Market1501~\cite{Market1501} and MSMT17~\cite{MSMT17}. Meanwhile, as the number of parameters increases, the performance of each model consistently improves. These results further indicate that our CION can be effectively used to pre-train various models, enabling them to achieve significant performance improvements for person ReID.

\end{document}